\documentclass[conference]{IEEEtran}
\usepackage[utf8]{inputenc}
\usepackage{cite}
\usepackage{amsmath,amssymb,amsfonts}
\usepackage{algorithmic}
\usepackage{graphicx}
\usepackage{textcomp}
\usepackage{url}
\usepackage{xcolor}
\usepackage{xspace}
\usepackage{rotating}
\usepackage{multirow}
\usepackage{tabularx}
\def\BibTeX{{\rm B\kern-.05em{\sc i\kern-.025em b}\kern-.08em
    T\kern-.1667em\lower.7ex\hbox{E}\kern-.125emX}}

\newcommand{\iamondb}{IAM-OnDB\xspace}

\newcommand{\num}[1]{#1\xspace} 

\newlength{\blackoutwidth}
\newcommand{\blackout}[1]
{%
  \settowidth{\blackoutwidth}{#1}%
  \rule[-0.3em]{\blackoutwidth}{1.125em}%
}
\renewcommand{\blackout}[1]{#1}

\usepackage{array}
\newcolumntype{H}{>{\setbox0=\hbox\bgroup}c<{\egroup}@{}}
\newcolumntype{Z}{>{\setbox0=\hbox\bgroup}c<{\egroup}@{\hspace*{-\tabcolsep}}}

\begin{document}

\title{IndyLSTMs: Independently Recurrent LSTMs}
\author{\IEEEauthorblockN{Pedro Gonnet, Thomas Deselaers}
\IEEEauthorblockA{\textit{Google AI Perception} \\
Zurich, Switzerland \\
{\tt [gonnet,deselaers]@google.com}}
}

\maketitle
\IEEEpeerreviewmaketitle

\begin{abstract}
    We introduce Independently Recurrent Long Short-term Memory cells: IndyLSTMs.
    These differ from regular LSTM cells in that the recurrent weights are not modeled as a full matrix, but as a diagonal matrix, i.e.\ the output and state of each LSTM cell depends on the inputs and its own output/state, as opposed to the input and the outputs/states of all the cells in the layer.
    The number of parameters per IndyLSTM layer, and thus the number of FLOPS per evaluation, is linear in the number of nodes in the layer, as opposed to quadratic for regular LSTM layers, resulting in potentially both smaller and faster models.

    We evaluate their performance experimentally by training several models on the popular \iamondb and CASIA online handwriting datasets, as well as on several of our in-house datasets.
    We show that IndyLSTMs, despite their smaller size, consistently outperform regular LSTMs both in terms of accuracy per parameter, and in best accuracy overall.
    We attribute this improved performance to the IndyLSTMs being less prone to overfitting.
\end{abstract}


\section{Introduction}

In this paper we introduce Independently Recurrent Long-Short-term Memory cells, i.e.\ \emph{IndyLSTMs}.
IndyLSTMs are a variation on Long Short-term Memory (LSTM) Cell neural networks  \cite{Hochreiter1997LSTM} where individual units within a hidden layer are not interconnected across time-steps -- inspired by the IndRNN architecture \cite{indrnn}.
We are using IndyLSTMs for the task of \emph{Online Handwriting Recognition}, where given a user input in the form of an \emph{ink}, i.e.\ a sequence of touch points, the task is to output the textual interpretation of this input.

Online handwriting recognition is gaining importance for two main reasons:
(a) Stylus-based input is gaining importance on tablet devices, e.g.\ the iPad Pro\footnote{\url{https://www.apple.com/ipad-pro/}}, Microsoft Surface devices\footnote{\url{https://www.microsoft.com/en-us/store/b/surface}}, and Chromebooks with styluses\footnote{\url{https://store.google.com/product/google_pixelbook}};
(b) an increasing number of users world-wide obtaining access to computing devices, many of whom are using scripts that are not as easy to type as English and thus handwriting is a useful means to enter text in their native language and script.

Furthermore, online handwriting recognition systems ideally run on the users' computing device and thus efficiency in terms of memory consumption and compute power are important features.
We show experimentally that the proposed IndyLSTM models are not only smaller and faster but also are able to obtain better recognition results because they are easier to train and less prone to overfitting.

\noindent\textbf{Related work:} Long short-term memory (LSTM)-based neural networks have become the \emph{de facto} standard for sequence modelling because they are easy to train and give good results \cite{jozefowicz2015empirical}.
Several other recurrent network cells have been proposed \cite{DBLP:journals/corr/ChoMGBSB14,chen2017minimalrnn,laurent2016recurrent,zoph2016neural,BradburyQRNN} and more recently, the idea of an IndRNN \cite{indrnn} has been proposed.
The IndRNN cell is based on the standard RNN cell, but removes the connection between different cells in the same layer from one layer to the next.
We base this work on the same ideas as the IndRNN work, but apply it to the more commonly used LSTM cell.

\section{IndyLSTMs}
\label{sec:indylstms}

The updates in regular Recurrent Neural Networks (RNN) \cite{jordan1997serial} can be described as:
\begin{equation}
    \mathbf h_{t} = \sigma\left(\mathbf W \mathbf x_t + \mathbf U \mathbf h_{t-1} + \mathbf b\right)
    \label{eqn:rnn}
\end{equation}
\noindent where $\mathbf x_t$ and $\mathbf h_t$ are vector representing the input and hidden state of the network at time $t$, respectively, $\sigma$ is an activation function, e.g. $\tanh$, and the matrices $\mathbf W$ and $\mathbf U$, as well as the bias vector $\mathbf b$, are parameters of the RNN layer.

Equation~\ref{eqn:rnn} can be interpreted as every element of the output/hidden state $\mathbf h_t$ being a non-linear function of a linear combination of all the inputs $\mathbf x_t$, plus a linear combination of all the previous outputs/hidden states $\mathbf h_{t-1}$.

Similarly, the updates to IndRNNs are described as:
\begin{equation}
    \mathbf h_t = \sigma\left(\mathbf W \mathbf x_t + \mathbf u \odot \mathbf h_{t-1} + \mathbf b\right)
    \label{eqn:indrnn}
\end{equation}
\noindent where $\mathbf u \odot \mathbf h_t$ is the elementwise product of the coefficient vector $\mathbf u$ with the previous outputs/hidden states $\mathbf h_{t-1}$, which replaces the matrix-vector product $\mathbf U \mathbf h_{t-1}$ in Equation~\ref{eqn:rnn}.
This is equivalent to requiring the matrix $\mathbf U$ to be diagonal.

Equation~\ref{eqn:indrnn} can be interpreted as every element of the ouput/hidden state $\mathbf h_t$ being a non-linear function of a linear combination of all the inputs $\mathbf x_t$, plus the element's own previous output/hidden state in $\mathbf h_{t-1}$.
In other words, each output relies only on it's own previous state, and not on that of the other states in the same layer.

Similarly to the approch in IndRNNs, we start with the updates in a Long-Short Term Memory (LSTM) layer \cite{Hochreiter1997LSTM}:
\begin{eqnarray}
    \mathbf f_t &=& \sigma_g(\mathbf W_{f} \mathbf x_t + \mathbf U_{f} \mathbf h_{t-1} + \mathbf b_f) \label{eqn:lstm} \\
    \mathbf i_t &=& \sigma_g(\mathbf W_{i} \mathbf x_t + \mathbf U_{i} \mathbf h_{t-1} + \mathbf b_i) \nonumber \\
    \mathbf o_t &=& \sigma_g(\mathbf W_{o} \mathbf x_t + \mathbf U_{o} \mathbf h_{t-1} + \mathbf b_o) \nonumber \\
    \mathbf c_t &=& \mathbf f_t \odot \mathbf c_{t-1} + \mathbf i_t \odot \sigma_c(\mathbf W_{c} \mathbf x_t + \mathbf U_{c} \mathbf h_{t-1} + \mathbf b_c) \nonumber \\
    \mathbf h_t &=& \mathbf o_t \odot \sigma_h(\mathbf c_t) \nonumber
\end{eqnarray}
\noindent where $\mathbf c_t$ is the cell state vector, and $\mathbf f_t$, $\mathbf i_t$, and $\mathbf o_t$ are the forget, input, and output activation vectors, respectively, and $\sigma_{[g|c|h]}$ are activation functions.
For inputs of length $n$ and $m$ cells, the matrices $\mathbf W_{[f|i|o|c]}$ of size $m\times n$, and $\mathbf U_{[f|i|o|c]}$ of size $m\times m$, along with the bias vectors $\mathbf b_{[f|i|o|c]}$ of length $m$, are the parameters of the LSTM layer.
Similarly to the RNN, each element of output/hidden state depends on all the entries of the input vector $\mathbf x_t$, as well as all the entries of the previous outputs and hidden states in $\mathbf h_{t-1}$ and $\mathbf c_{t-1}$, respectively.\footnote{Note that although the cell states $\mathbf c_{t-1}$ are only used element-wise in each update, they are used to compute the outputs $\mathbf h_{t-1}$, which are used across the entire layer.}

We then re-write Equation~\ref{eqn:lstm} as:
\begin{eqnarray}
    \mathbf f_t &=& \sigma_g(\mathbf W_{f} \mathbf x_t + \mathbf u_{f} \odot \mathbf h_{t-1} + \mathbf b_f) \label{eqn:indylstm} \\
    \mathbf i_t &=& \sigma_g(\mathbf W_{i} \mathbf x_t + \mathbf u_{i} \odot \mathbf h_{t-1} + \mathbf b_i) \nonumber \\
    \mathbf o_t &=& \sigma_g(\mathbf W_{o} \mathbf x_t + \mathbf u_{o} \odot \mathbf h_{t-1} + \mathbf b_o) \nonumber \\
    \mathbf c_t &=& \mathbf f_t \odot \mathbf c_{t-1} + \mathbf i_t \odot \sigma_c(\mathbf W_{c} \mathbf x_t + \mathbf u_{c} \odot \mathbf h_{t-1} + \mathbf b_c) \nonumber \\
    \mathbf h_t &=& \mathbf o_t \odot \sigma_h(\mathbf c_t) \nonumber
\end{eqnarray}
\noindent where we replace the matrix-vector products $\mathbf U_{[f|i|o|c]} \mathbf h_{t-1}$ with the element-wise products $\mathbf u_{[f|i|o|c]} \odot \mathbf h_{t-1}$, where $\mathbf u_{[f|i|o|c]}$ are vectors of length $m$.
As with the IndRNNs, each element of output/hidden state depends on all the entries of the input vector $\mathbf x_t$, as well as only its own previous output and hidden state $\mathbf h_{t-1}$ and $\mathbf c_{t-1}$, respectively.
In other words, each output relies only on it's own state, and not on that of all the states in the layer.
We refer to Equation~\ref{eqn:indylstm} as \emph{Independent Long Short-Term Memory} (IndyLSTM) units.\footnote{We chose the name IndyLSTM over IndLSTM, and other potentially more precise abbreviations, simply because we were teenagers in the nineties, and Grunge fans to this day, and so to us it sounds cooler.}

For an LSTM layer with $n$ inputs and $m$ outputs, the number of parameters is $4mn + 4m^2 + 4m = 4m(n + m + 1)$, which are the sizes of the matrices $\mathbf W_{[f|i|o|c]}$ and $\mathbf U_{[f|i|o|c]}$, and the bias vectors $\mathbf b_{[f|i|o|c]}$, respectively.
For an IndyLSTM layer with $n$ inputs and $m$ outputs, the number of parameters is $4mn + 4m + 4m = 4m(n + 2)$, which are the size of the matrices $\mathbf W_{[f|i|o|c]}$, and the vectors $\mathbf u_{[f|i|o|c]}$ and $\mathbf b_{[f|i|o|c]}$, respectively.
Since the computational cost of evaluating Equations~\ref{eqn:lstm} and \ref{eqn:indylstm} are linear in the number of parameters, i.e. each parameter corresponds roughly to a single multiply-add operation, the cost of evaluating a single step in an IndyLSTM layer is roughly $n/(n+m)$ times smaller than for regular LSTMs\footnote{Our implementation of IndyLSTM is available as \texttt{\footnotesize tf.contrib.rnn.IndyLSTMCell} in TensorFlow 1.10.0~\cite{tensorflow2015-whitepaper}.}.

In the following sections, we quantify how IndyLSTMs compare to regular LSTMs in terms of model accuracy over size, and by extension speed, and in terms of accuracy overall.

\section{Experimental setup}

\begin{figure}
    \centering
    \includegraphics[width=0.8\linewidth]{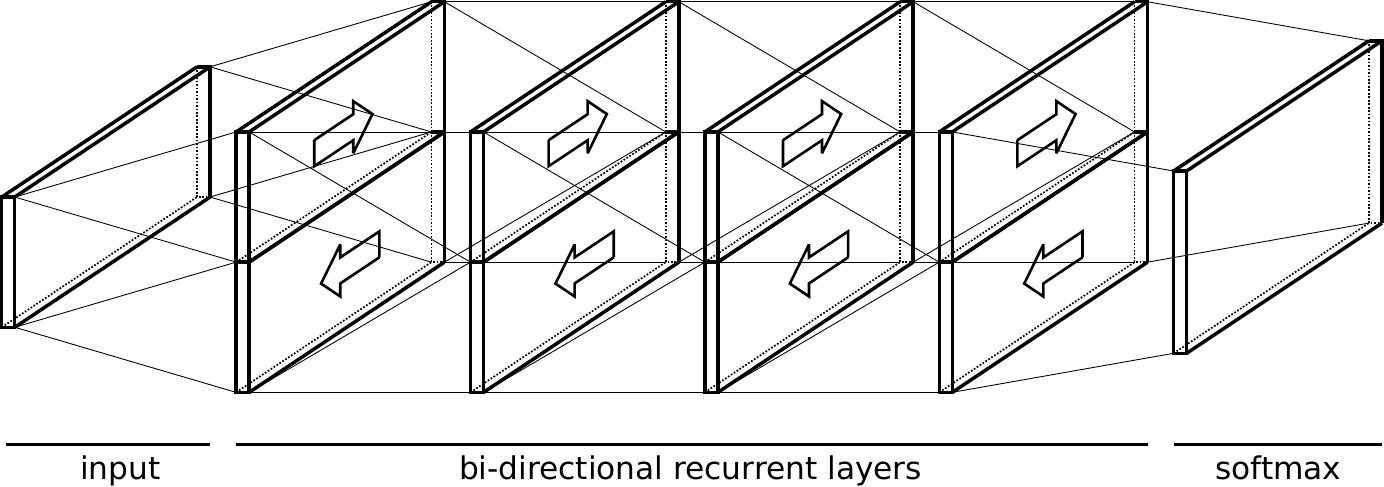}
    \caption{Schema of the model architecture used in this paper, i.e. a sequence of inputs is passed through 3--9 bi-directional recurrent layers of the same size, and finally through a softmax layer.}
    \label{fig:model}
\end{figure}

We use online handwriting recognition \cite{vcarbune2019:hwr} as a model problem to compare IndyLSTMs to regular LSTMs, using publicly available datasets.

The end-to-end model architecture we use is the same as the one described in \cite{vcarbune2019:hwr}, i.e. one or more layers of bi-directional LSTM/IndyLSTM of the same width, followed by a single softmax layer.
We convert the handwriting input, which is a sequence of raw touch points, to a sequence of B\'ezier curves as described in \cite{vcarbune2019:hwr} and use them as inputs to the model, and Connectionist Temporal Classification (CTC) \cite{Graves2006ConnectionistTC} to convert the sequence of softmax outputs to a sequence of characters.

The models are trained with a batch size of 8, on 10 parallel replicas, using the Adam Optimizer \cite{Kingma2014AdamAM} with a fixed learning rate of $0.0001$.
All weight matrices are initialized using the Glorot uniform initializer \cite{pmlr-v9-glorot10a}, all bias vectors were initialized to zero, and the IndyLSTM weights $\mathbf u_{[f|i|o|c]}$ were initialized with uniform random values in $[-1,1]$.
Random dropout \cite{pham2014dropout} is applied to the outputs of each LSTM/IndyLSTM layer.
Training is stopped after 5\,M steps without improvement on the validation data.

\noindent For each dataset, we train models varying:
\begin{itemize}
    \item The recurrent bi-directional layer type, using either LSTMs or IndyLSTMs,
    \item The number of layers in $\{3, \dots, 9\}$,
    \item The width of the layers in $\{32, \dots, 256\}$, in increments of $32$,
    \item The dropout rate on the LSTM/IndyLSTM layer outputs during training in $\{0, \dots, 0.6\}$ in increments of $0.1$, and up to $0.7$ in cases where the best results were obtained with $0.6$.
\end{itemize}
and compare them in terms of Character Error Rate (CER) versus model size.

Although some of the models might have benefited from varying a number of other hyper-parameters, we choose the current simple setup for the sake of experimental clarity.

The bulk of our experiments were run on the \iamondb and CASIA datasets described below.

\subsection{\iamondb}

The \iamondb dataset \cite{liwicki:icdar05} is the most used evaluation dataset for online handwriting recognition.
It consists of 298 523 characters in 86\,272 word instances from a dictionary of 11\,059 words using 79 different characters written by 221 writers.
We use the standard \iamondb dataset separation: one training set and a test set containing 5\,363 and 3\,859 written lines, respectively.
The reported CER is the error rate computed over the test set.

\subsection{CASIA}

The ICDAR-2013 Competition for Online Handwriting Chinese Character Recognition \cite{yin2013icdar} introduced a dataset for classifying the most common Chinese characters. 
Following the experimental setup in \cite{casiaRNNnet4}, the dataset used for training is the CASIA database \cite{CASIA-data2011}, which contains 2\,693\,183 samples for training and 224\,590 samples for testing, produced by different writers.
The number of character classes is 3\,755.

Note that although our model is fully capable of recognizing character sequences, the CASIA training and test data consists of only single characters.

\subsection{In-house data and models}
\label{sec:in-house}
In order to compare our results to those presented in \cite{vcarbune2019:hwr}, we also train several models using our in-house datasets, which consist of data collected through prompting, commercially available data, artificially inflated data, and labeled/self-labeled anonymized recognition requests (see \cite{Google:HWRPAMI} for a more detailed description).

For each language, we train IndyLSTM models with 3--9 layers each, where the layer widths are chosen such that the resulting number of parameters is at most equal to that of the original LSTM-based models they are compared against.
E.g.\ since the production Latin recognizer in \cite{vcarbune2019:hwr} is a 5$\times$224 LSTM-based model with 5\,378\,088 parameters\footnote{This value is slightly larger than the 5\,281\,061 reported in \cite{vcarbune2019:hwr} since we have, since then, slightly increased the size of the recognized Latin alphabet to accommodate new languages.}, we train IndyLSTM-based models with 3$\times$398, 4$\times$327, 5$\times$284, 6$\times$254, 7$\times$232, 8$\times$215, and 9$\times$201 configurations.
As with the other experiments, we vary the dropout rate in the range $\{0, \dots, 0.5\}$ in increments of $0.1$.

\section{Results and discussion}

\subsection{\iamondb}

\begin{figure*}
    \centering
    \includegraphics[width=0.75\linewidth]{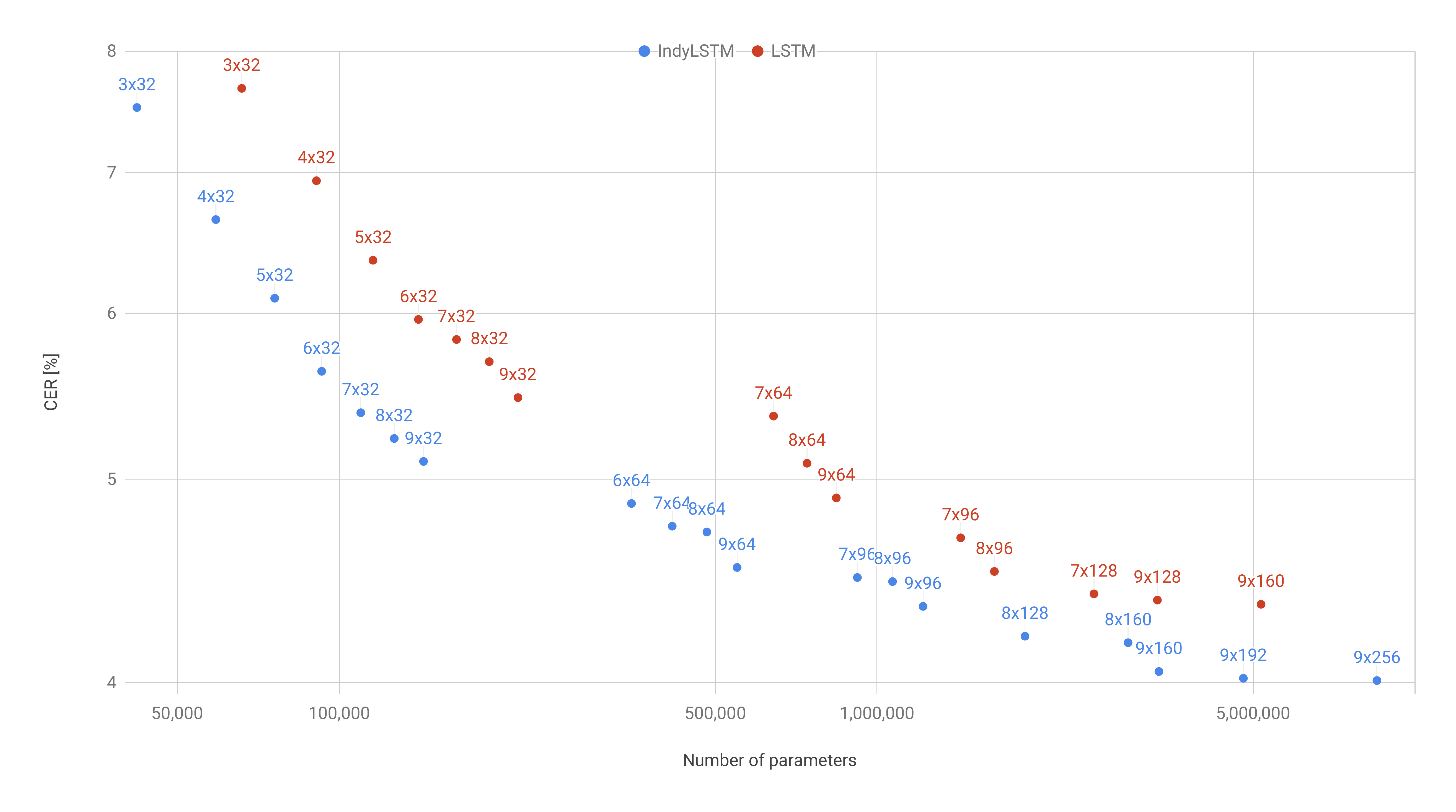}
    \includegraphics[width=0.75\linewidth]{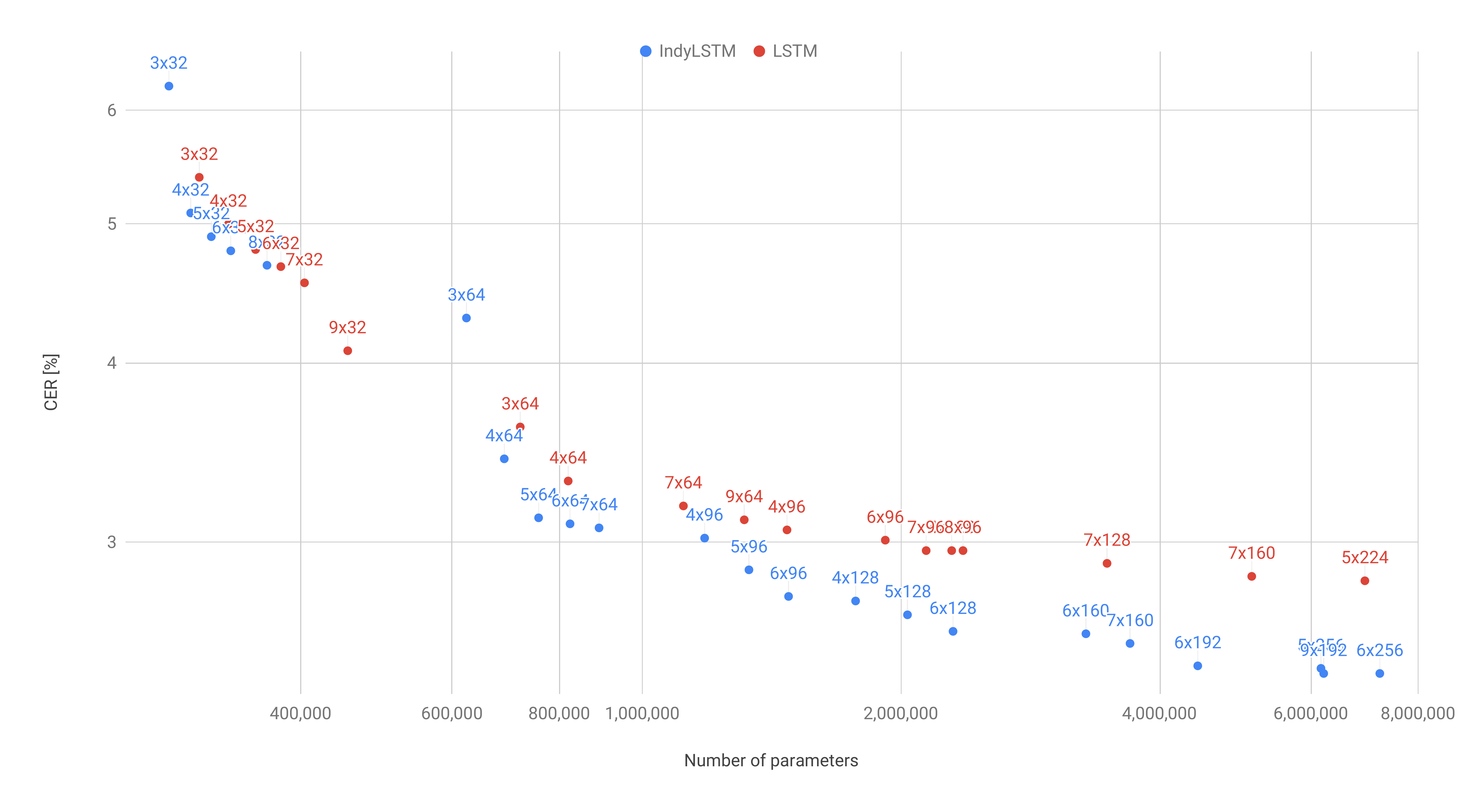}
    \caption{Character Error Rate (CER) of different models trained on the \iamondb (top) and CASIA (bottom) data. Each dot represents the best model for a given maximum number of parameters, i.e.\ the Pareto-optimal sets from the LSTM and IndyLSTM results over the different model depths, widths, and dropout rates used during training.}
\label{fig:joined}
\end{figure*}

Figure~\ref{fig:joined} shows the results for the best models, in terms of CER over the respective test set, per number of parameters trained on the \iamondb and CASIA datasets.
The full set of results can be seen in Appendix~\ref{sec:details}.

On the \iamondb datasets, over all model widths and depths, the IndyLSTMs out-performed the LSTMs with respect to CER vs.\ number of parameters, as well as in terms of accuracy overall.
Additionally, the IndyLSTMs consistently out-perform the LSTMs with respect to CER vs.\ depth and width, i.e. the 9$\times$160 IndyLSTM model with lowest CER out-performs the 9$\times$160 LSTM model with lowest CER, despite the former having only two thirds as many parameters as the latter.

The best-performing model, with a CER of 4.01\%, is the 9$\times$256 IndyLSTM (8.5\,M parameters), i.e.\ the deepest and widest architecture we tested, trained with a dropout rate of 0.6.
This beats the previously reported comparable state-of-the-art result in \cite{vcarbune2019:hwr} using a 5$\times$64 LSTM and obtaining a CER of 5.9\% without the use of language models and other feature functions.
The best 5$\times$64 IndyLSTM in this work obtains a CER of 5.15\%, thus also beating the previous state-of-the-art.

For comparison, we also evaluated the best-performing IndyLSTM model with the feature functions described in \cite{vcarbune2019:hwr}, namely

\begin{itemize}
    \item {\bf Character Language Model:}
    A 7-gram language model over Unicode codepoints from a large web-mined text corpus using stupid back-off~\cite{brants:prodlm2007} and pruned to \num{10} million \num{7}-grams. 
    
    \item {\bf Word Language Model:}
    We also use a word-based language model trained on a similar corpus~\cite{speech-lm-mining,speech-lm-infra}, using 3-grams pruned to \num{1.25} million entries.
    
    \item {\bf Character Classes:}
    We add a scoring heuristic which boosts the score of characters from the language's alphabet, e.g.\ numbers, upper-case letters, punctuation, etc. 
\end{itemize}
\noindent The feature function weights are trained using the Google Vizier service and its default algorithm, specifically batched Gaussian process bandits, and expected improvement as the acquisition function \cite{golovin-vizier}. 
We run 7 Vizier studies, each performing 500 individual trials, and then pick the configuration that performed best across all trials.
The resulting recognizer obtains a CER of 2.91\%, which is significantly better than the current state of the art using feature functions of 4.0\% in \cite{vcarbune2019:hwr}.

\subsection{CASIA}

Over almost the entire range of model sizes (Figure~\ref{fig:joined}, bottom), the IndyLSTMs outperform the LSTMs in terms of accuracy over number of parameters, as well as in terms of accuracy overall.
The 6$\times$256 IndyLSTM model, trained with a dropout rate of $0.4$, obtained the lowest CER of 2.43\%.
This is 14\% better than the best CER of 2.82\% obtained by the 5$\times$224 LSTM-based model (6.9\,M parameters), trained with a dropout rate of $0.4$.

The best IndyLSTM models still fails to beat the current state of the art CER of 1.95\% in \cite{casiaRNNnet4} obtained with an ensemble model with 20.5M parameters and input sub-sequence sampling.
The latter model, however, used several data augmentation techniques for training and inference that were not used in our own experiments.

\subsection{In-house data and models}

\begin{table}
    \caption{Character Error Rate (CER) of best models trained using our in-house datasets for the languages in \cite{vcarbune2019:hwr}. The first pair of columns show the CER reported and model architecture used therein. The final two columns contain the CER of the best IndyLSTM model over different depth$\times$width combinations corresponding to roughly the same number of parameters as the original LSTM model, and over different dropout rates.}
    \label{tab:prod}
    \centering
    \begin{tabular}{|l|c|c|c|c|}
    \hline
        &  \multicolumn{2}{c|}{LSTM in \cite{vcarbune2019:hwr}} & \multicolumn{2}{c|}{best IndyLSTM} \\
        Language &  CER [\%] & model & CER [\%] & model \\ \hline\hline
        \scriptsize\sf en & 6.48  & \multirow{3}{*}{5$\times$224 (0.5)} & \bf 6.25 & \multirow{3}{*}{9$\times$201 (0.3)} \\ \cline{1-2}\cline{4-4}
        \scriptsize\sf de & 5.40  & & \bf 4.97 & \\ \cline{1-2}\cline{4-4}
        \scriptsize\sf es & \bf 4.56  & & 4.69 & \\ \hline
        \scriptsize\sf ar & 7.87  & 5$\times$160 (0.5) & \bf 6.7 & 6$\times$181 (0.3) \\ \hline
        \scriptsize\sf ko & 6.79 & 5$\times$160 (0.5) & \bf 6.65 & 9$\times$146 (0.2) \\ \hline
        \scriptsize\sf th & 1.78 & 5$\times$128 (0.5) & \bf 1.6 & 5$\times$162 (0.4) \\ \hline
        \scriptsize\sf hi & 7.42 & 5$\times$192 (0.5) & \bf 7.14 & 6$\times$218 (0.5) \\ \hline
        \scriptsize\sf zh & 1.39 & 4$\times$192 (0.5) & \bf 1.27 & 9$\times$166 (0.3) \\ \hline
    \end{tabular}
\end{table}

Following the procedure in Section~\ref{sec:in-house}, we train IndyLSTM-based models of different depths and widths for English ({\footnotesize\sf en}), German ({\footnotesize\sf de}), Spanish ({\footnotesize\sf es}), Arabic ({\footnotesize\sf ar}), Korean ({\footnotesize\sf ko}), Thai ({\footnotesize\sf th}), Devanagari ({\footnotesize\sf hi}), and Chinese ({\footnotesize\sf zh}).
Table~\ref{tab:prod} compares the best IndyLSTM models, in terms of CER over our test sets, with the corresponding results in \cite{vcarbune2019:hwr}, which were all trained with a dropout rate of $0.5$, as well as the best LSTM-based models re-trained varying the dropout rate.

In all cases, the best IndyLSTM models out-perform the original LSTM models, despite having at most as many parameters.
The best IndyLSTM models are neither consistently the deepest nor the widest, and the optimal dropout rate varies from model to model.

\begin{figure}
    \centering
    \includegraphics[width=\linewidth]{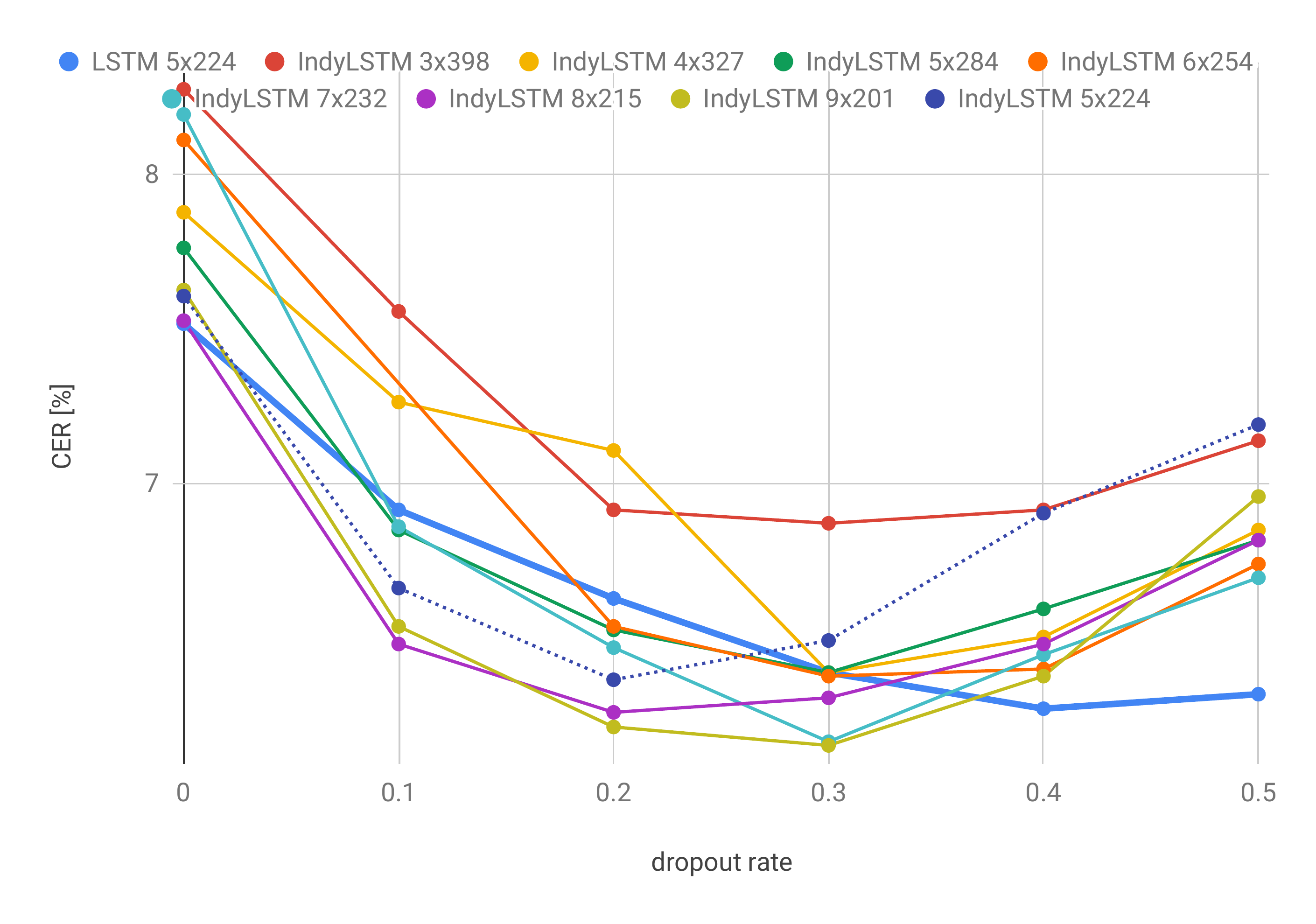}
    \caption{Character Error Rate (CER) over dropout rate for our LSTM-based production Latin model and different IndyLSTM-based models with 3 to 9 layers with the maximum width such that the number of parameters is less than that of the 5$\times$224 LSTM-based model, plus a 5$\times$224 IndyLSTM-based model for comparison. The IndyLSTM-based models with 7, 8, and 9 layers out-perform the LSTM-based model.}
    \label{fig:latin_prod}
\end{figure}

Figure~\ref{fig:latin_prod} shows in detail the CER for the English-language ({\footnotesize\sf en}) LSTM and IndyLSTM models.
With the exception of the 3-layer IndyLSTM model, all models perform more or less in the same range of accuracy, and the differences between one model and another are smaller than the differences due to the different dropout rates within the same model.

\subsection{Speed}

On the \iamondb data, over all model sizes and dropout ratios, the IndyLSTMs required roughly 1.8 more epochs to \textit{train} than the regular LSTM models of the same depth and width.
For the CASIA data set, and our own data sets, the number of training steps was roughly equal over all model sizes and dropout ratios.

In order to verify how the size differences affect the \textit{inference} speed, we implemented the IndyLSTM layer in TensorFlow \cite{tensorflow2015-whitepaper} analogously to the {\small\sf BasicLSTMLayer} therein and ran the models trained over the \iamondb dataset through TensorFlow's {\small\sf benchmark\_model} tool.
Over all model sizes, the IndyLSTMs were roughly 20\% faster than their LSTM counterparts, on average, despite containing roughly 37\% fewer parameters, on average.
We attribute this difference in performance to overheads within TensorFlow, e.g.\ the 3$\times$128 LSTM model spends less than 22\% of the total time in the {\small\sf MatMul} op, which constitutes the bulk of the computation, while the {\small\sf Merge}, {\small\sf Const} and {\small\sf Select} ops make up for 24\% of the total time.

\section{Conclusions}

We introduce IndyLSTMs, an independently-recurrent variant of the well-known LSTM recursive neural networks, and compare their accuracy on both academic datasets, and our in-house production datasets.

Our experiments show that, over a wide range of online handwriting recognition tasks, IndyLSTMs outperform the LSTMs in terms of accuracy over number of parameters, as well as in terms of accuracy overall.
In many cases, IndyLSTM models outperform LSTM models with the same depth and width, and thus $\sim$50\% more parameters.

Furthermore, the IndyLSTM models presented herein establish new state of the art results on the \iamondb dataset, both with and without the use of feature functions.
The results are not just restricted academic datasets such as \iamondb and CASIA; on our in-house datasets, the IndyLSTM models consistently out-performed their production LSTM counterparts.
Due to their better accuracy, smaller size, and faster execution, IndyLSTMs are interesting models for on-device inference, where resource consumption is often a limiting factor.

\begin{figure}
    \centering
    \includegraphics[width=\linewidth]{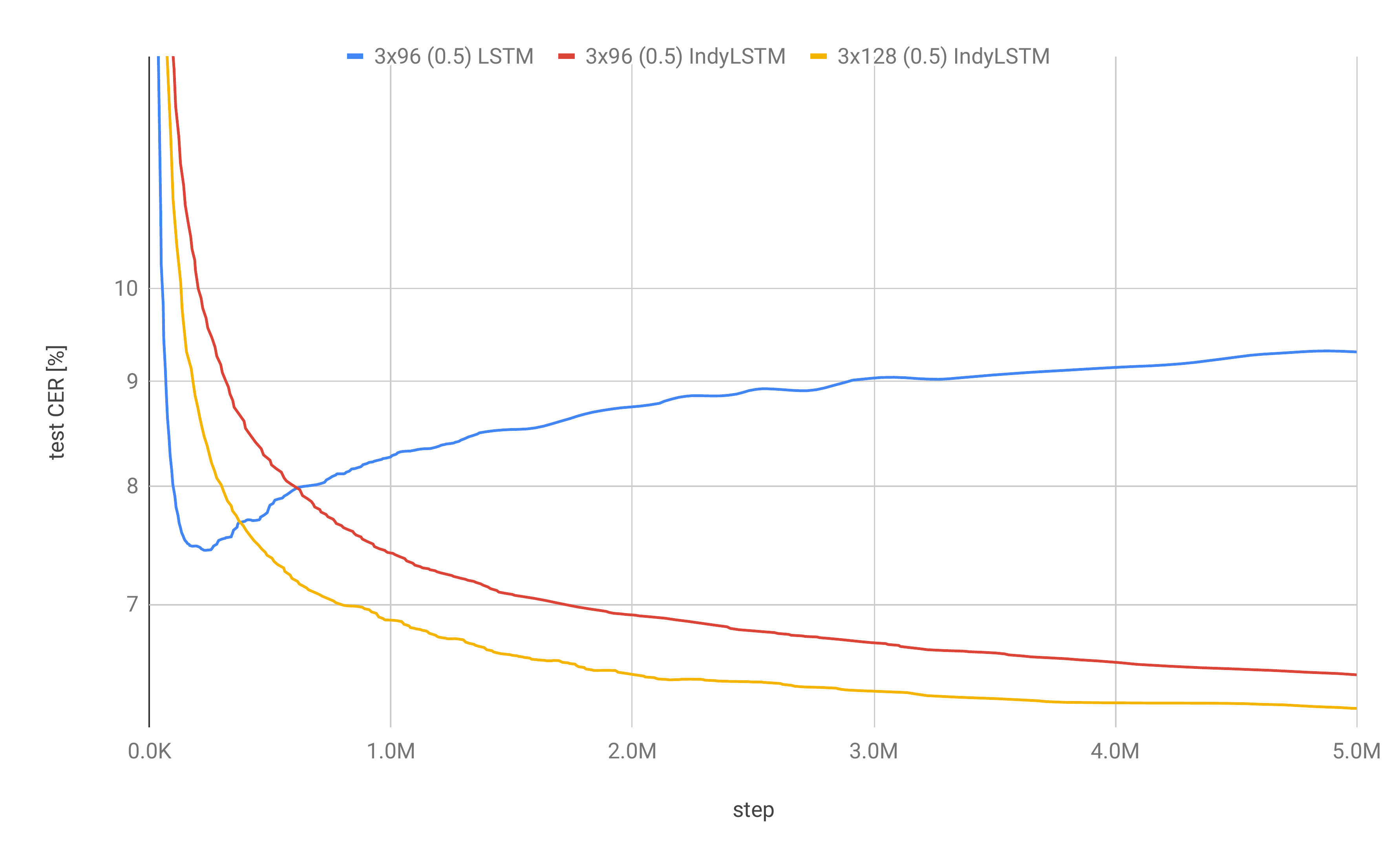}
    \includegraphics[width=\linewidth]{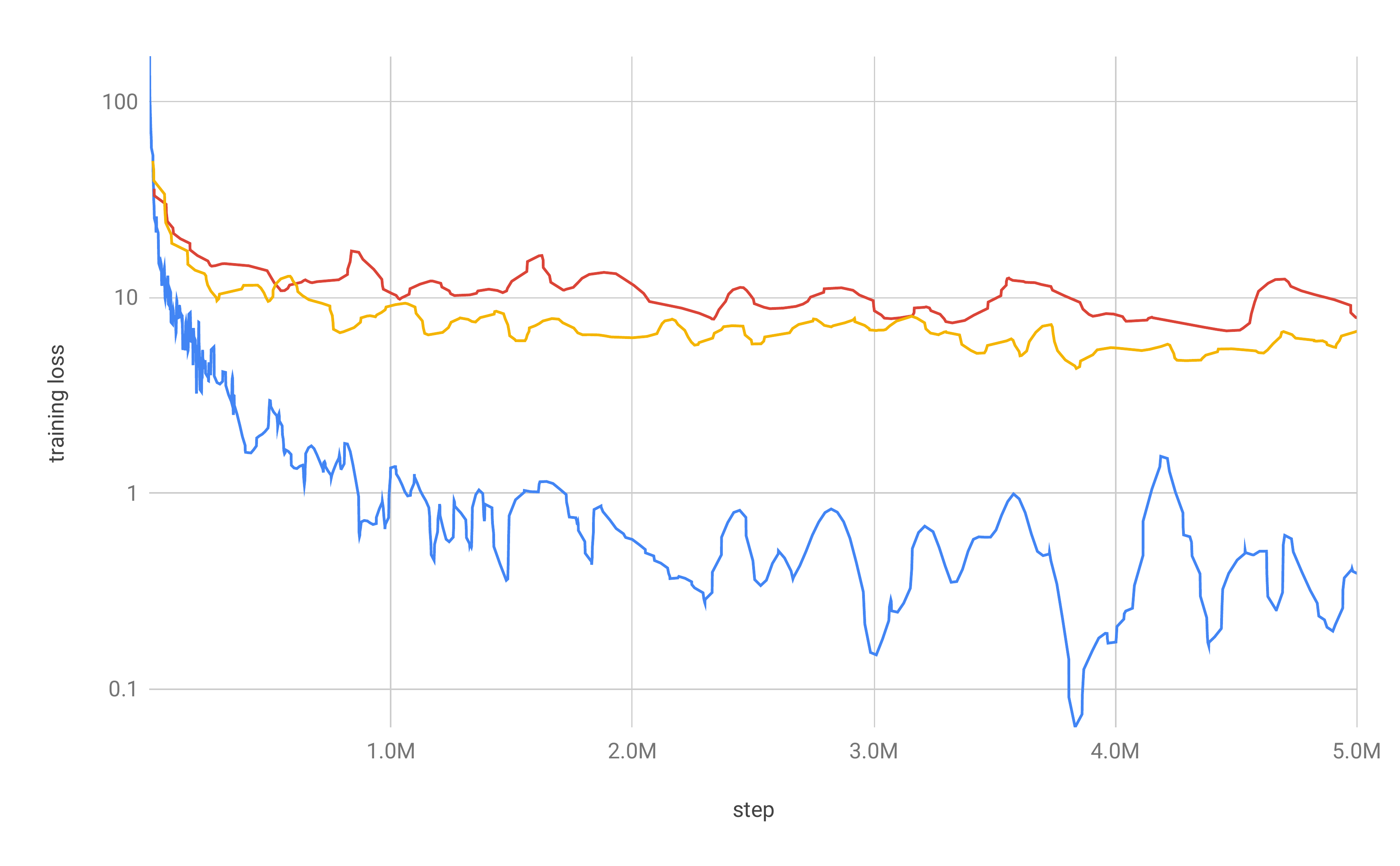}
    \caption{Character Error Rate (CER) computed over the \iamondb test set during training (top) and training loss (below) of the 3$\times$96 LSTM model (541\,520 parameters), and the 3$\times$96 and 3$\times$128 IndyLSTM models (322\,640 and 561\,232 parameters, respectively). While the LSTM model overfits after roughly 200k steps, i.e.\ the test error grows while the training loss essentially vanishes, both IndyLSTM models converge more or less smoothly.}
    \label{fig:overfit}
\end{figure}

We attribute this improved accuracy to IndyLSTMs being \textit{less prone to overfitting} than regular LSTMs.
Figure~\ref{fig:overfit} shows the evolution of the CER computed over the \iamondb test data and the training loss for the 3$\times$96 LSTM model (541\,520 parameters), and the 3$\times$96 and 3$\times$128 IndyLSTM models (322\,640 and 561\,232 parameters, respectively). 
The 3$\times$128 IndyLSTM was chosen since it has roughly as many parameters as the 3$\times$96 LSTM.
All three models were trained with a dropout rate of 0.5, which produced the best result for each architecture.
After $\sim$200k steps, the LSTM model's training loss stops decreasing and starts increasing while the training loss all but vanishes, i.e.\ the model overfits to the training data.
The IndyLSTM models, on the other hand, do not seem to overfit, i.e.\ the test error decreases monotonically while the training loss levels off after $\sim$1M steps.

The overfitting hypothesis is also supported by comparing our in-house English-language data results to the \iamondb results.
The datasets differ in that the former is more than 10$\times$ larger than the latter.\footnote{Although our in-house data supports a larger alphabet (313 symbols, all characters for all Latin script-based languages), the bulk of the data consist of the same 79 symbols used in the \iamondb data.}
Given the order of magnitude size difference, overfitting should be much less likely to occur with our in-house training datasets than with the \iamondb dataset, and the advantage of using IndyLSTMs over LSTMs should be smaller.
This is effectively the case: on the \iamondb data, IndyLSTMs consistently out-perform their LSTM counterparts for the same architecture and models with roughly the same number of parameters differ by 5--20\% (see Figure~\ref{fig:joined}), whereas on our in-house data, the 5$\times$224 IndyLSTM model does \textit{not} out-perform the 5$\times$224 LSTM model (see Figure~\ref{fig:latin_prod}), and the (best) 9$\times$201 IndyLSTM model with the same number of parameters is only 3.5\% better than the LSTM model (see Table~\ref{tab:prod}).

Unfortunately, at this point we can only speculate as to \textit{why} the IndyLSTMs are less prone to overfitting than LSTMs, even for models with the same number of parameters.

We hypothesize that it is the difference in model complexity that make the IndyLSTM less likely to overfit than LSTMs.
The number of parameters is, in this case, a poor measure for model complexity, since as shown in Figure~\ref{fig:overfit}, the 3$\times$128 IndyLSTM has slightly more parameters than the 3$\times$96 LSTM, but does not overfit.
We therefore assume that it is the \textit{number of connections} which make the difference: as pointed out in Section~\ref{sec:indylstms}, the output of every node in an IndyLSTM layer depends only on the input and the node's previous output and hidden state, whereas the output every node in an LSTM layer depends on the input and the previous output and hidden state of every other node in the layer.
These additional connections allow the LSTM layers to much better adapt to the data, which, in the use cases studied herein, only leads to overfitting.

\section*{Acknowledgments}

The authors would like to thank \blackout{Victor Carbune}, \blackout{Henry Rowley}, and \blackout{Yasuhisa Fuji} for their valuable input on both the methods presented here, as well as on the manuscript itself.

\bibliographystyle{IEEEtran}
\bibliography{hwr}

\begin{thebibliography}{10}
\providecommand{\url}[1]{#1}
\csname url@samestyle\endcsname
\providecommand{\newblock}{\relax}
\providecommand{\bibinfo}[2]{#2}
\providecommand{\BIBentrySTDinterwordspacing}{\spaceskip=0pt\relax}
\providecommand{\BIBentryALTinterwordstretchfactor}{4}
\providecommand{\BIBentryALTinterwordspacing}{\spaceskip=\fontdimen2\font plus
\BIBentryALTinterwordstretchfactor\fontdimen3\font minus
  \fontdimen4\font\relax}
\providecommand{\BIBforeignlanguage}[2]{{%
\expandafter\ifx\csname l@#1\endcsname\relax
\typeout{** WARNING: IEEEtran.bst: No hyphenation pattern has been}%
\typeout{** loaded for the language `#1'. Using the pattern for}%
\typeout{** the default language instead.}%
\else
\language=\csname l@#1\endcsname
\fi
#2}}
\providecommand{\BIBdecl}{\relax}
\BIBdecl

\bibitem{Hochreiter1997LSTM}
S.~Hochreiter and J.~Schmidhuber, ``Long short-term memory,'' \emph{Neural
  Computation}, vol.~9, no.~8, pp. 1735--1780, Nov. 1997.

\bibitem{indrnn}
S.~Li, W.~Li, C.~Cook, C.~Zhu, and Y.~Gao, ``Independently recurrent neural
  network (indrnn): Building a longer and deeper rnn,'' in \emph{CVPR}, 2018.

\bibitem{jozefowicz2015empirical}
R.~Jozefowicz, W.~Zaremba, and I.~Sutskever, ``An empirical exploration of
  recurrent network architectures,'' in \emph{International Conference on
  Machine Learning}, 2015, pp. 2342--2350.

\bibitem{DBLP:journals/corr/ChoMGBSB14}
\BIBentryALTinterwordspacing
K.~Cho, B.~van Merrienboer, {\c{C}}.~G{\"{u}}l{\c{c}}ehre, F.~Bougares,
  H.~Schwenk, and Y.~Bengio, ``Learning phrase representations using {RNN}
  encoder-decoder for statistical machine translation,'' \emph{CoRR}, vol.
  abs/1406.1078, 2014. [Online]. Available:
  \url{http://arxiv.org/abs/1406.1078}
\BIBentrySTDinterwordspacing

\bibitem{chen2017minimalrnn}
M.~Chen, ``Minimalrnn: Toward more interpretable and trainable recurrent neural
  networks,'' \emph{arXiv preprint arXiv:1711.06788}, 2017.

\bibitem{laurent2016recurrent}
T.~Laurent and J.~von Brecht, ``A recurrent neural network without chaos,''
  \emph{arXiv preprint arXiv:1612.06212}, 2016.

\bibitem{zoph2016neural}
B.~Zoph and Q.~V. Le, ``Neural architecture search with reinforcement
  learning,'' \emph{arXiv preprint arXiv:1611.01578}, 2016.

\bibitem{BradburyQRNN}
J.~Bradbury, S.~Merity, C.~Xiong, and R.~Socher, ``Quasi-recurrent neural
  networks,'' \emph{CoRR}, vol. abs/1611.01576, 2016.

\bibitem{jordan1997serial}
M.~I. Jordan, ``Serial order: A parallel distributed processing approach,'' in
  \emph{Advances in psychology}.\hskip 1em plus 0.5em minus 0.4em\relax
  Elsevier, 1997, vol. 121, pp. 471--495.

\bibitem{tensorflow2015-whitepaper}
\BIBentryALTinterwordspacing
M.~Abadi, A.~Agarwal, P.~Barham, E.~Brevdo, Z.~Chen, C.~Citro, G.~S. Corrado,
  A.~Davis, J.~Dean, M.~Devin, S.~Ghemawat, I.~Goodfellow, A.~Harp, G.~Irving,
  M.~Isard, Y.~Jia, R.~Jozefowicz, L.~Kaiser, M.~Kudlur, J.~Levenberg,
  D.~Man\'{e}, R.~Monga, S.~Moore, D.~Murray, C.~Olah, M.~Schuster, J.~Shlens,
  B.~Steiner, I.~Sutskever, K.~Talwar, P.~Tucker, V.~Vanhoucke, V.~Vasudevan,
  F.~Vi\'{e}gas, O.~Vinyals, P.~Warden, M.~Wattenberg, M.~Wicke, Y.~Yu, and
  X.~Zheng, ``{TensorFlow}: Large-scale machine learning on heterogeneous
  systems,'' 2015, software available from tensorflow.org. [Online]. Available:
  \url{https://www.tensorflow.org/}
\BIBentrySTDinterwordspacing

\bibitem{vcarbune2019:hwr}
V.~Carbune, P.~Gonnet, T.~Deselaers, H.~A. Rowley, A.~Daryin, M.~Calvo, L.-L.
  Wang, D.~Keysers, S.~Feuz, and P.~Gervais, ``Fast multi-language lstm-based
  online handwriting recognition,'' \emph{arXiv preprint arXiv:1902.10525},
  2019.

\bibitem{Graves2006ConnectionistTC}
A.~Graves, S.~Fern{\'a}ndez, F.~J. Gomez, and J.~Schmidhuber, ``Connectionist
  temporal classification: labelling unsegmented sequence data with recurrent
  neural networks,'' in \emph{ICML}, 2006.

\bibitem{Kingma2014AdamAM}
D.~P. Kingma and J.~Ba, ``Adam: A method for stochastic optimization,''
  \emph{ICLR}, 2014.

\bibitem{pmlr-v9-glorot10a}
X.~Glorot and Y.~Bengio, ``Understanding the difficulty of training deep
  feedforward neural networks,'' in \emph{Proceedings of the Thirteenth
  International Conference on Artificial Intelligence and Statistics},
  vol.~9.\hskip 1em plus 0.5em minus 0.4em\relax PMLR, 2010, pp. 249--256.

\bibitem{pham2014dropout}
V.~Pham, T.~Bluche, C.~Kermorvant, and J.~Louradour, ``Dropout improves
  recurrent neural networks for handwriting recognition,'' in \emph{Frontiers
  in Handwriting Recognition (ICFHR), 2014 14th International Conference
  on}.\hskip 1em plus 0.5em minus 0.4em\relax IEEE, 2014, pp. 285--290.

\bibitem{liwicki:icdar05}
M.~Liwicki and H.~Bunke, ``{IAM-OnDB}-an on-line {English} sentence database
  acquired from handwritten text on a whiteboard,'' in \emph{{ICDAR}}, 2005,
  pp. 956--961.

\bibitem{yin2013icdar}
F.~Yin, Q.-F. Wang, X.-Y. Zhang, and C.-L. Liu, ``Icdar 2013 {Chinese}
  handwriting recognition competition,'' in \emph{Document Analysis and
  Recognition (ICDAR), 2013 12th International Conference on}.\hskip 1em plus
  0.5em minus 0.4em\relax IEEE, 2013, pp. 1464--1470.

\bibitem{casiaRNNnet4}
X.~Zhang, F.~Yin, Y.~Zhang, C.~Liu, and Y.~Bengio, ``Drawing and recognizing
  {Chinese} characters with recurrent neural network,'' \emph{ArXiV}, 2016.

\bibitem{CASIA-data2011}
C.-L. Liu, F.~Yin, D.-H. Wang, and Q.-F. Wang, ``{CASIA} online and offline
  {Chinese} handwriting databases,'' in \emph{ICDAR}, 2011.

\bibitem{Google:HWRPAMI}
D.~Keysers, T.~Deselaers, H.~Rowley, L.-L. Wang, and V.~Carbune,
  ``Multi-language online handwriting recognition,'' \emph{{IEEE} Trans.
  Pattern Analysis \& Machine Intelligence}, vol.~39, no.~6, pp. 1180--1194,
  2017.

\bibitem{brants:prodlm2007}
T.~Brants, A.~C. Popat, P.~Xu, F.~J. Och, and J.~Dean, ``Large language models
  in machine translation,'' in \emph{{EMNLP-CoNLL}}, 2007, pp. 858--867.

\bibitem{speech-lm-mining}
M.~Prasad, T.~Breiner, and D.~van Esch, ``Mining training data for language
  modeling across the world’s languages,'' in \emph{Proceedings of the 6th
  International Workshop on Spoken Language Technologies for Under-resourced
  Languages (SLTU 2018)}, 2018.

\bibitem{speech-lm-infra}
M.~Chua, D.~van Esch, N.~Coccaro, E.~Cho, S.~Bhandari, and L.~Jia, ``Text
  normalization infrastructure that scales to hundreds of language varieties,''
  in \emph{Proceedings of the 11th edition of the Language Resources and
  Evaluation Conference}, 2018.

\bibitem{golovin-vizier}
D.~Golovin, B.~Solnik, S.~Moitra, G.~Kochanski, J.~E. Karro, and D.~Sculley,
  Eds., \emph{Google Vizier: A Service for Black-Box Optimization}, 2017.

\end{thebibliography}

\appendices

\section{Detailed results}
\label{sec:details}

\begin{sidewaysfigure*}
    \centering
    \includegraphics[width=0.49\linewidth]{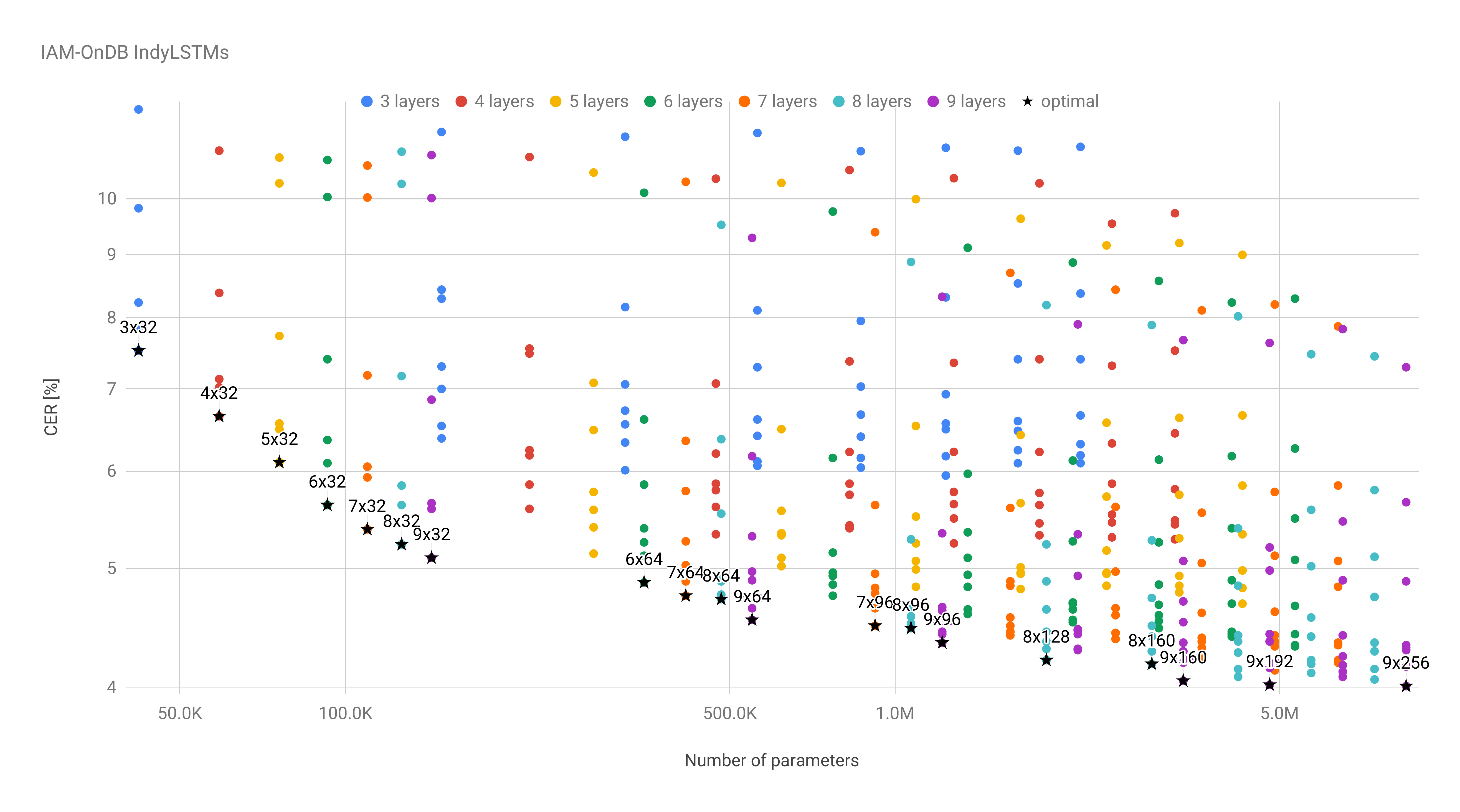}
    \includegraphics[width=0.49\linewidth]{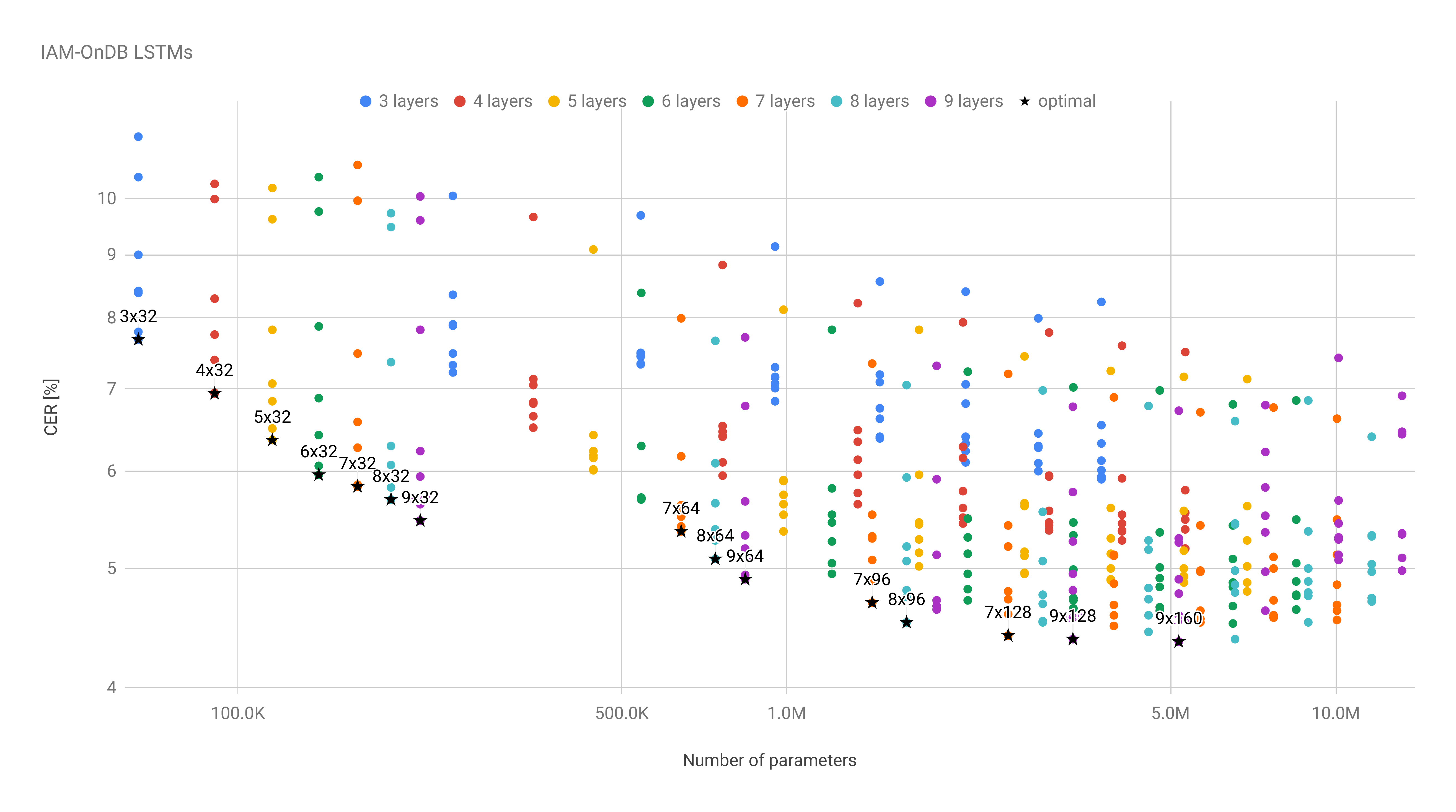}
    \includegraphics[width=0.49\linewidth]{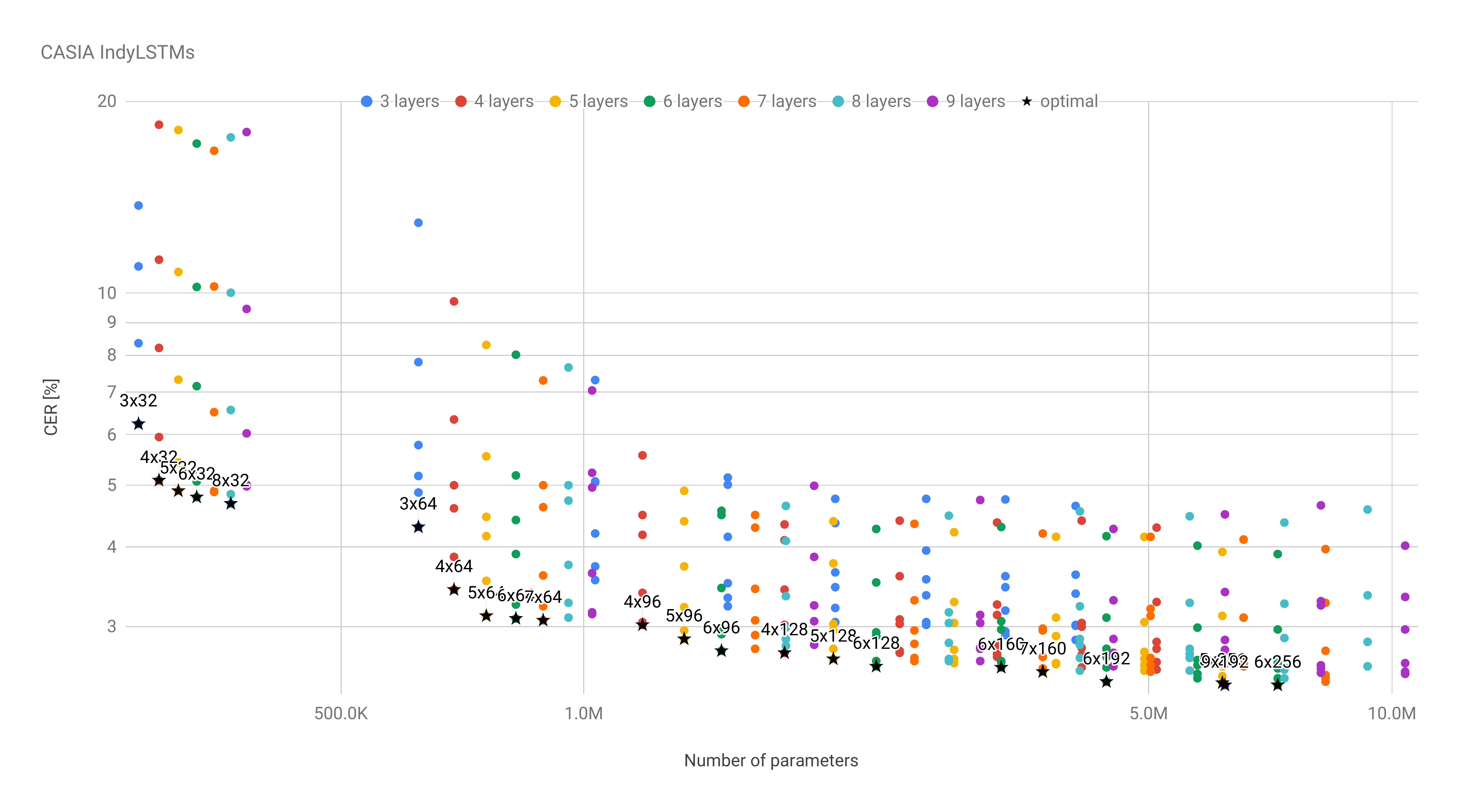}
    \includegraphics[width=0.49\linewidth]{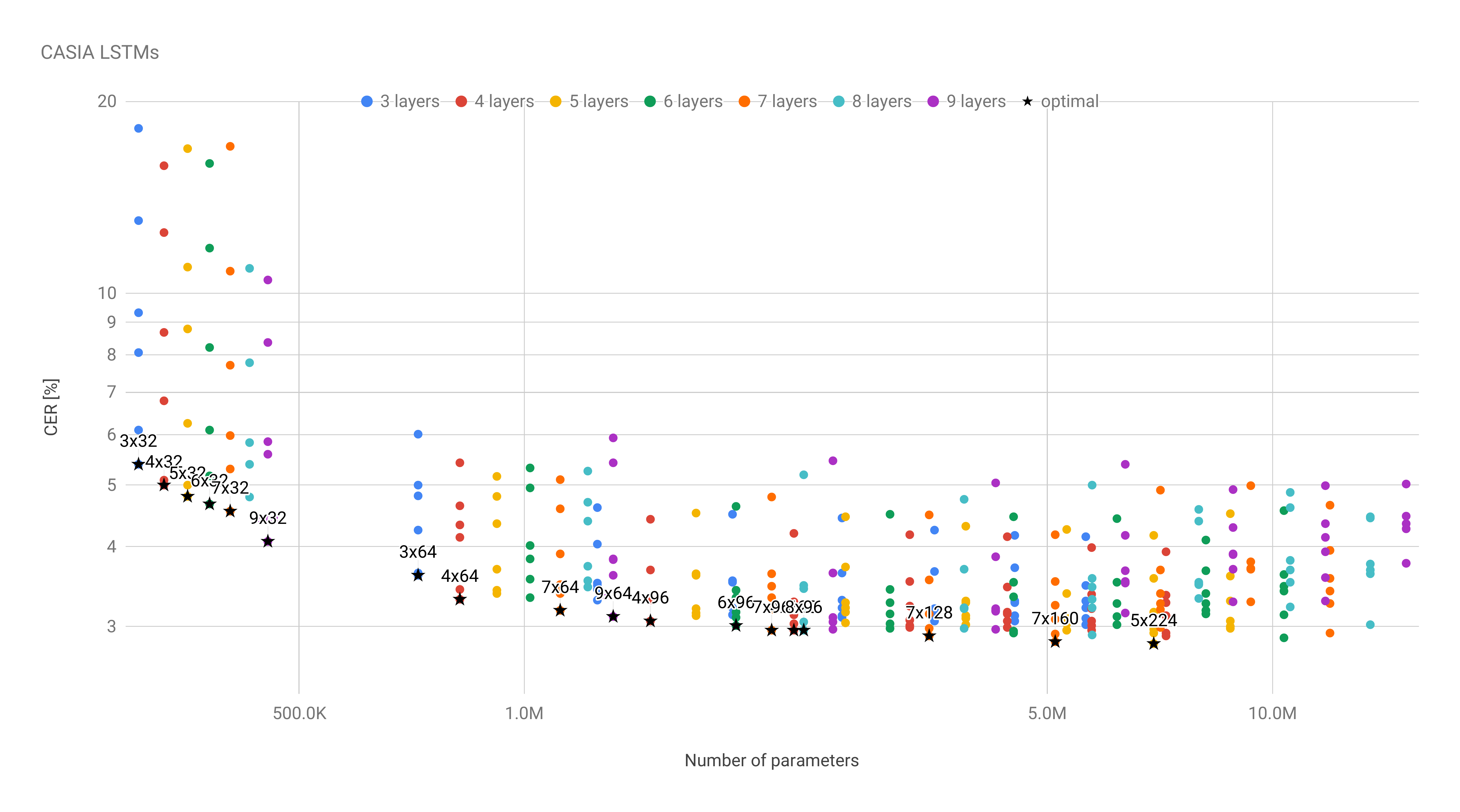}
    \caption{Character Error Rate (CER) of different models trained on the \iamondb (top) and CASIA (bottom) datasets. Each dot represents a single training run, and stars indicate the best model for a given maximum number of parameters, and different colours correspond to different numbers of layers. Vertically aligned dots of the same colour correspond to training runs of the same architecture with different dropout rates.}
\label{fig:full_results}
\end{sidewaysfigure*}

Figure~\ref{fig:full_results} shows the CER versus the number of parameters for each model trained on both the \iamondb and CASIA datasets.

\section{Optimal dropout rate}

\begin{figure*}
    \centering
    \includegraphics[width=0.31\linewidth]{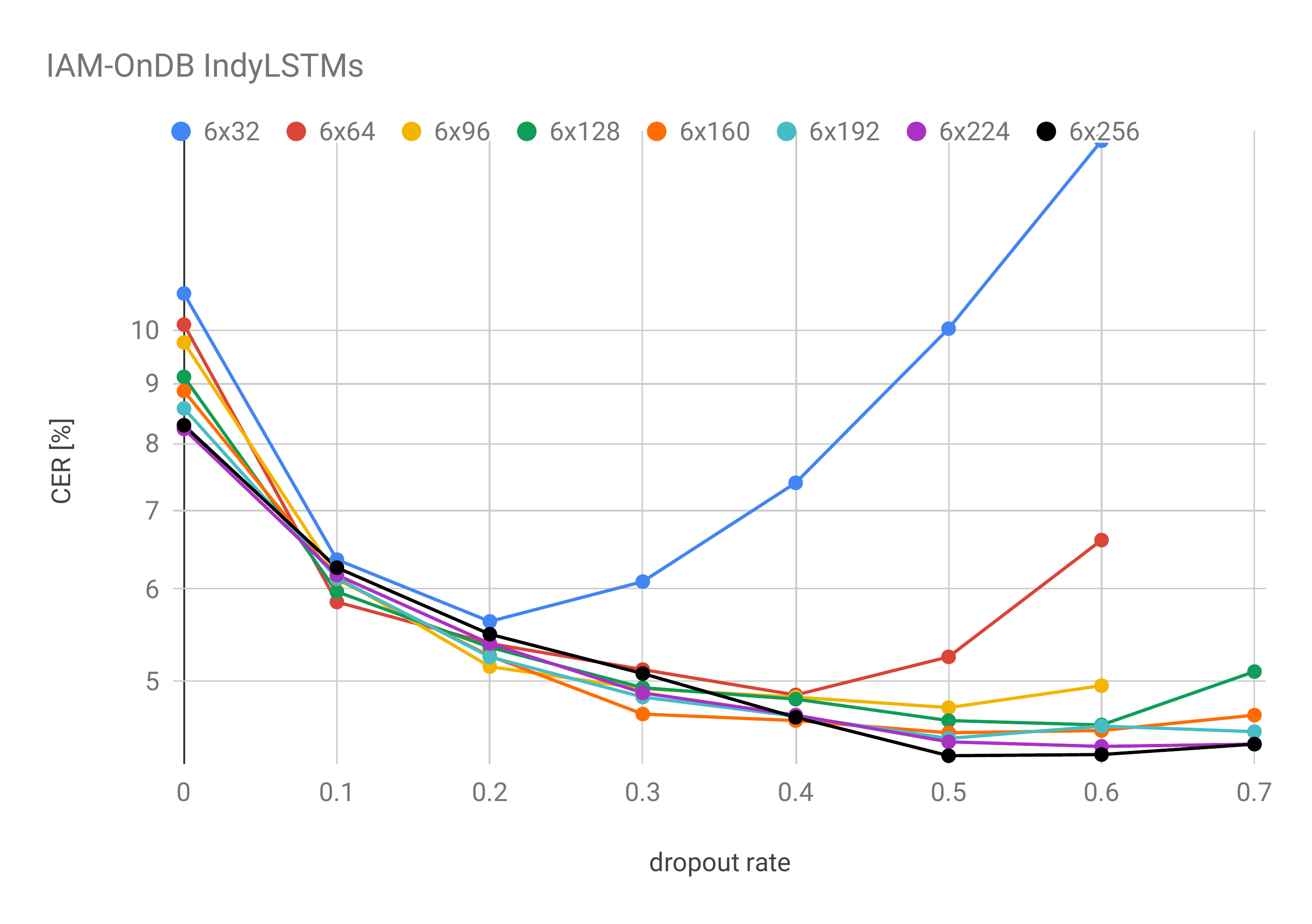}\hspace{3ex}
    \includegraphics[width=0.31\linewidth]{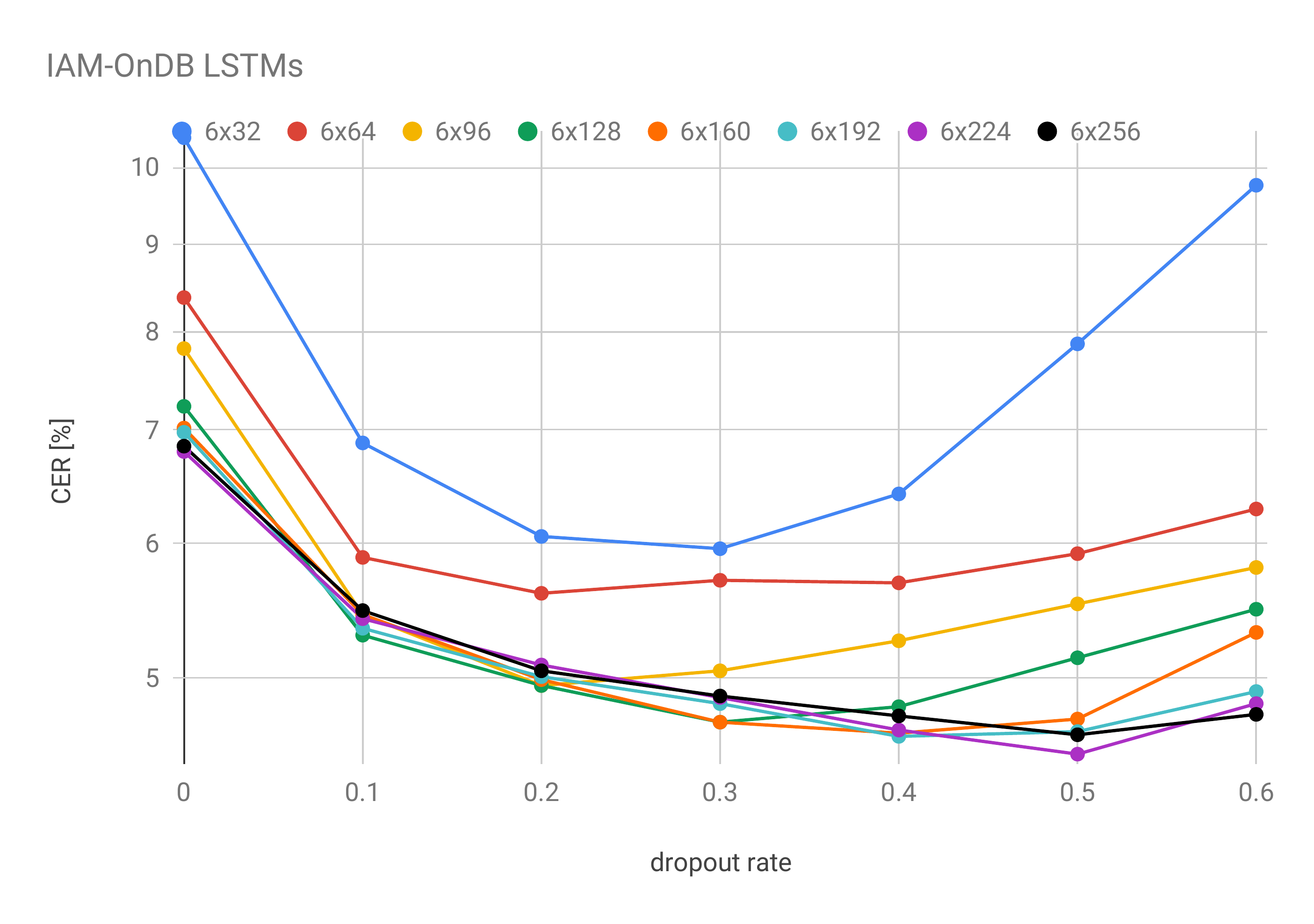} \\
    \includegraphics[width=0.31\linewidth]{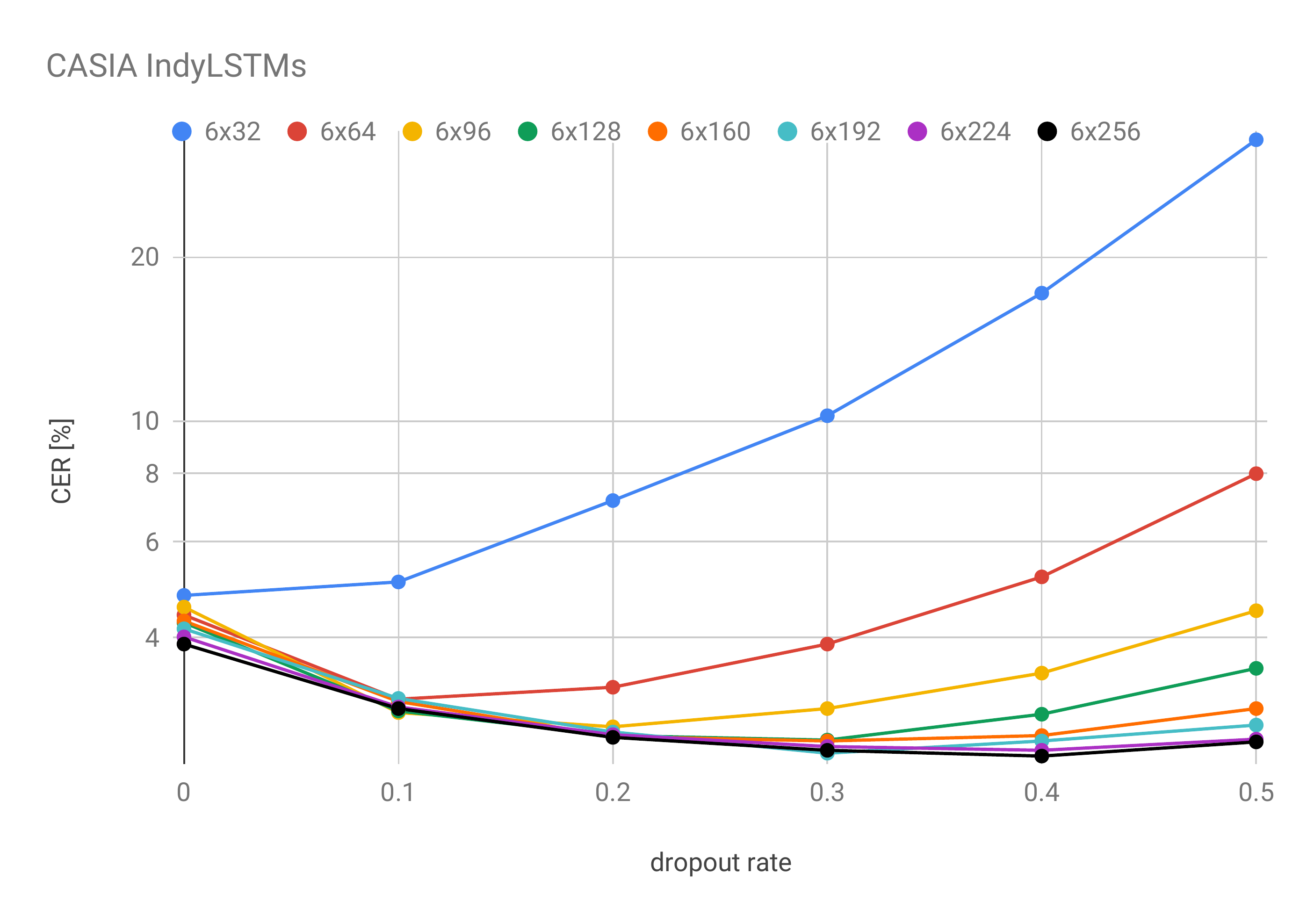}\hspace{3ex}
    \includegraphics[width=0.31\linewidth]{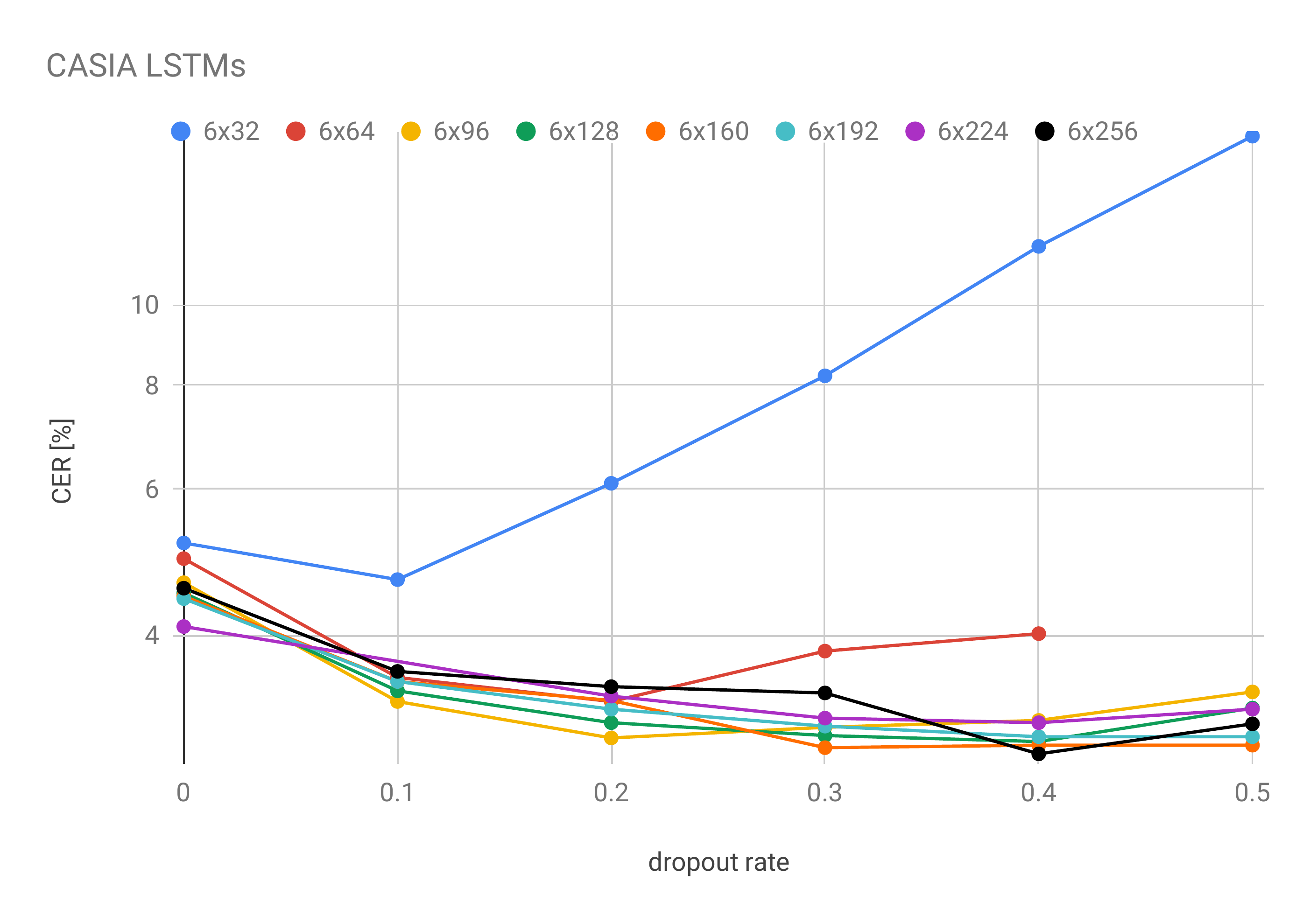}
    \caption{Character Error Rate (CER) over dropout rate as a function of model width for the \iamondb and CASIA datasets using IndyLSTMs and LSTMs. For both datasets and models, the optimal dropout rate increases with the model width.}
    \label{fig:dropout_width}
\end{figure*}

\begin{figure*}
    \centering
    \includegraphics[width=0.31\linewidth]{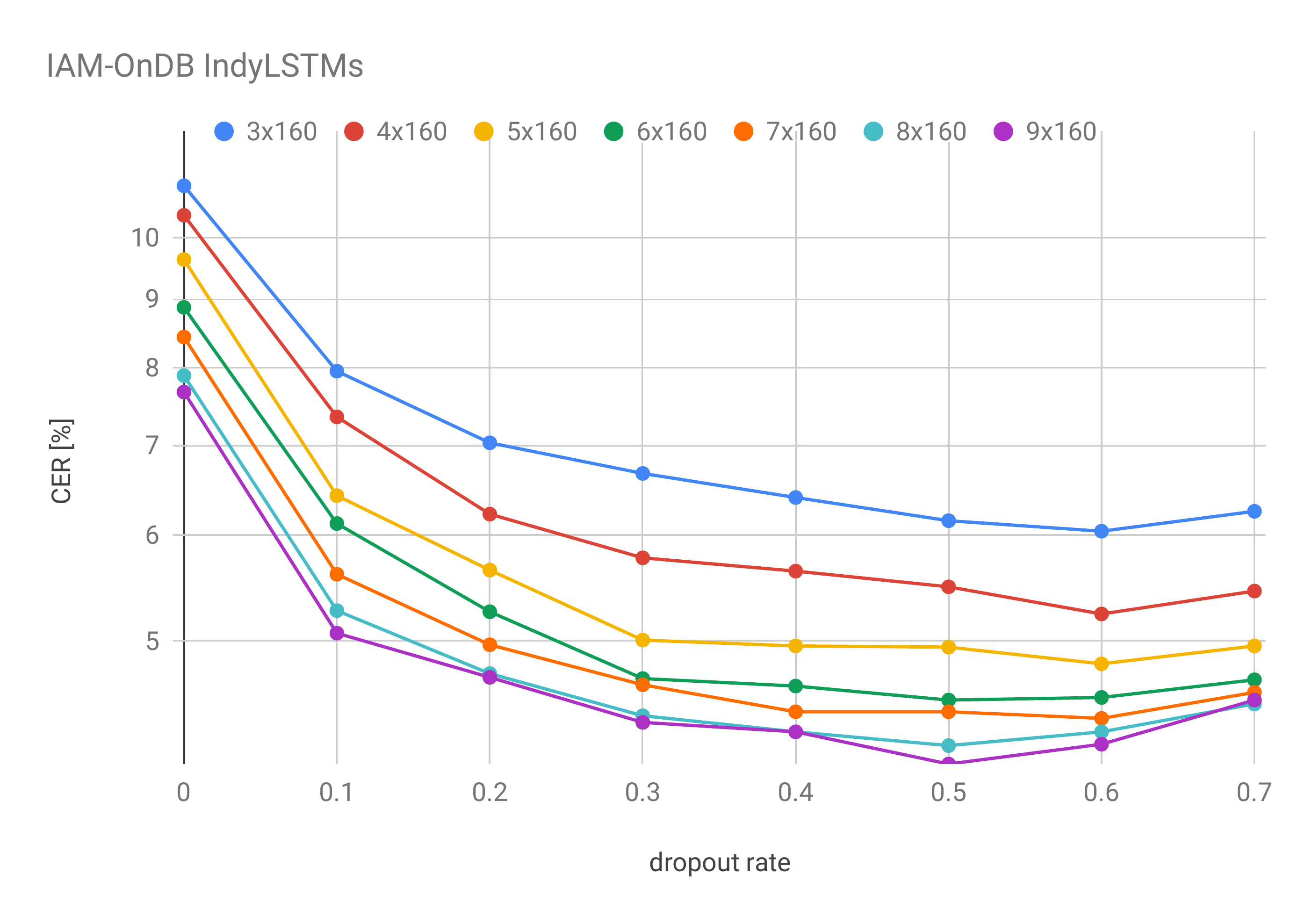}\hspace{3ex}
    \includegraphics[width=0.31\linewidth]{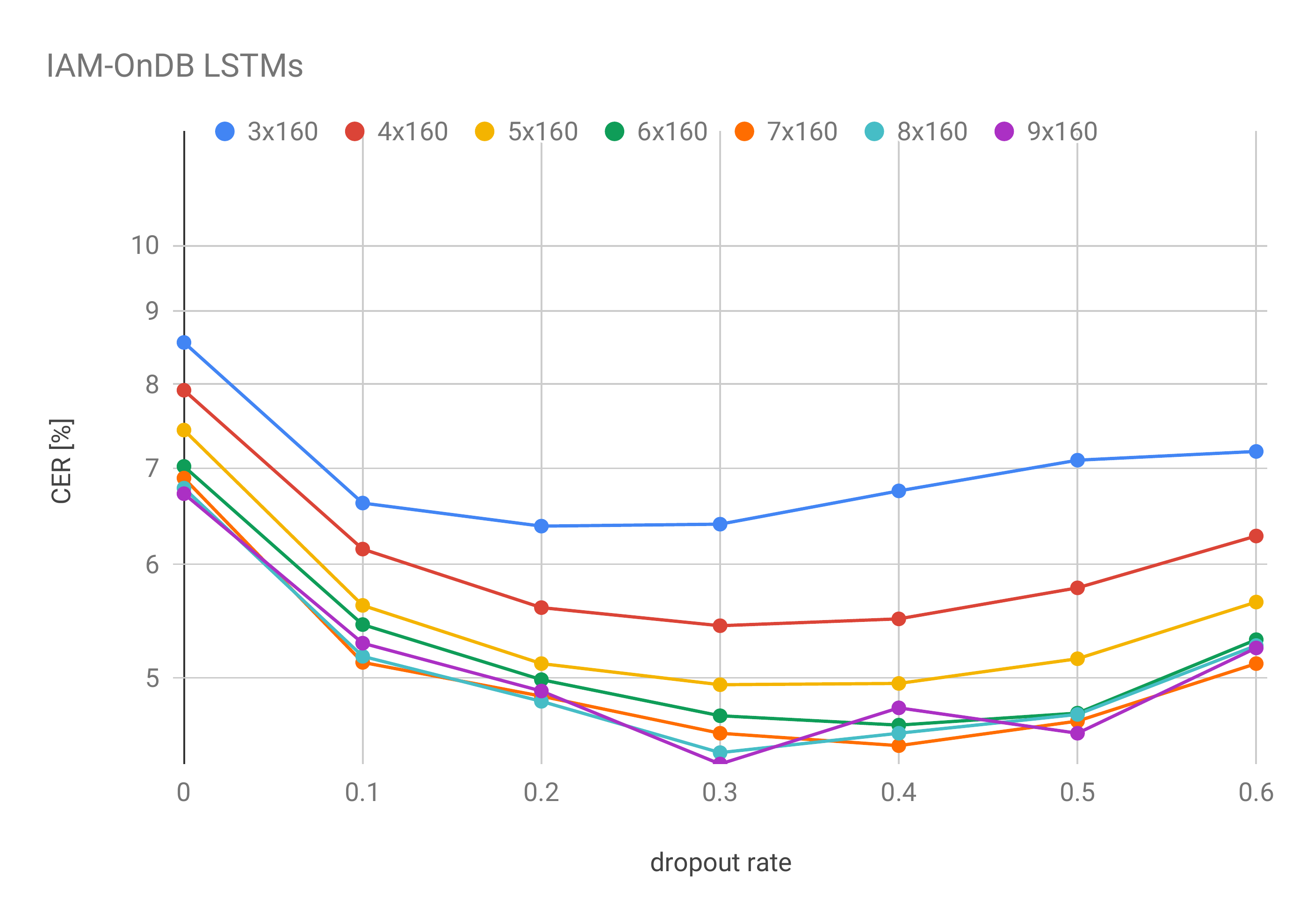} \\
    \includegraphics[width=0.31\linewidth]{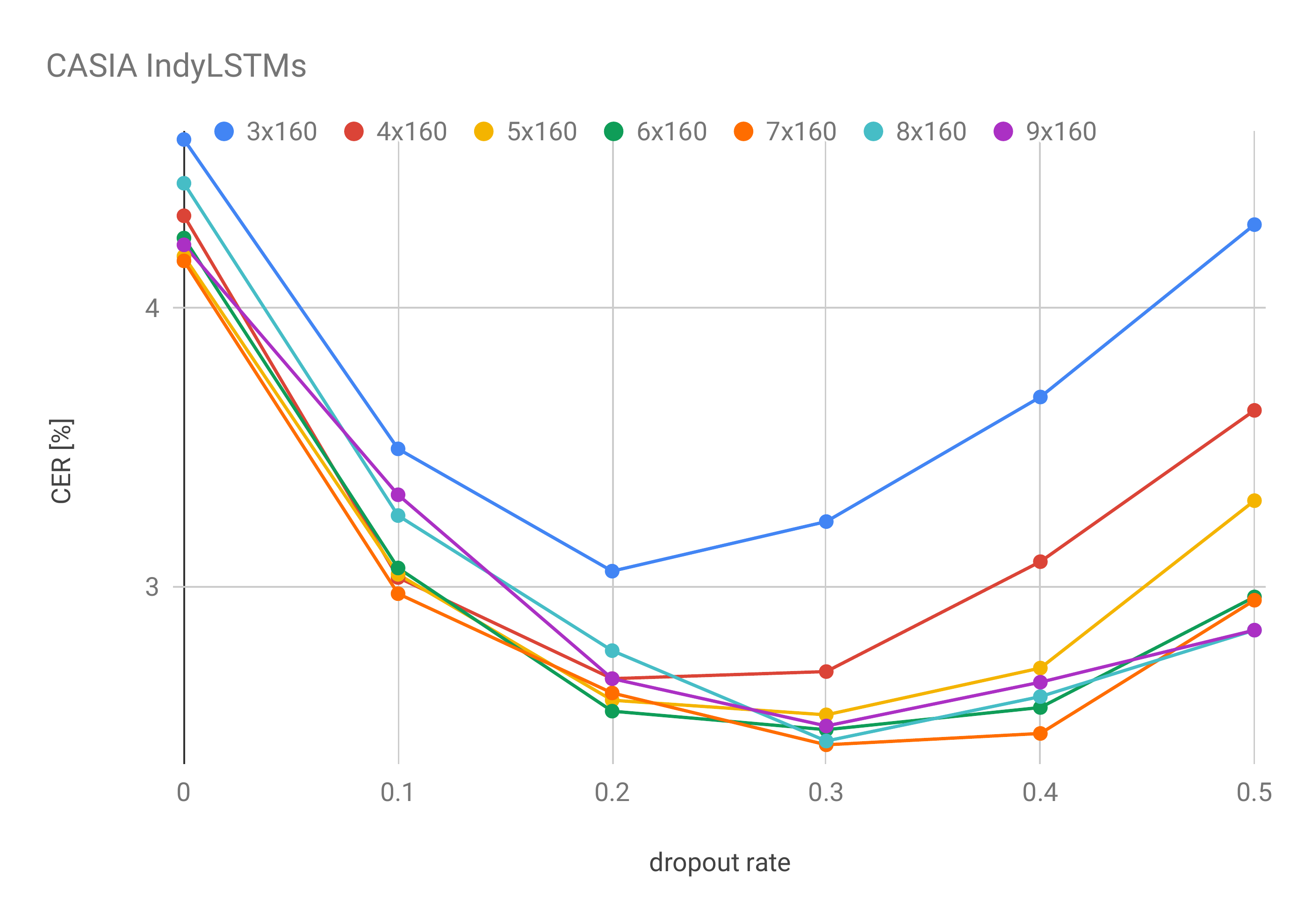}\hspace{3ex}
    \includegraphics[width=0.31\linewidth]{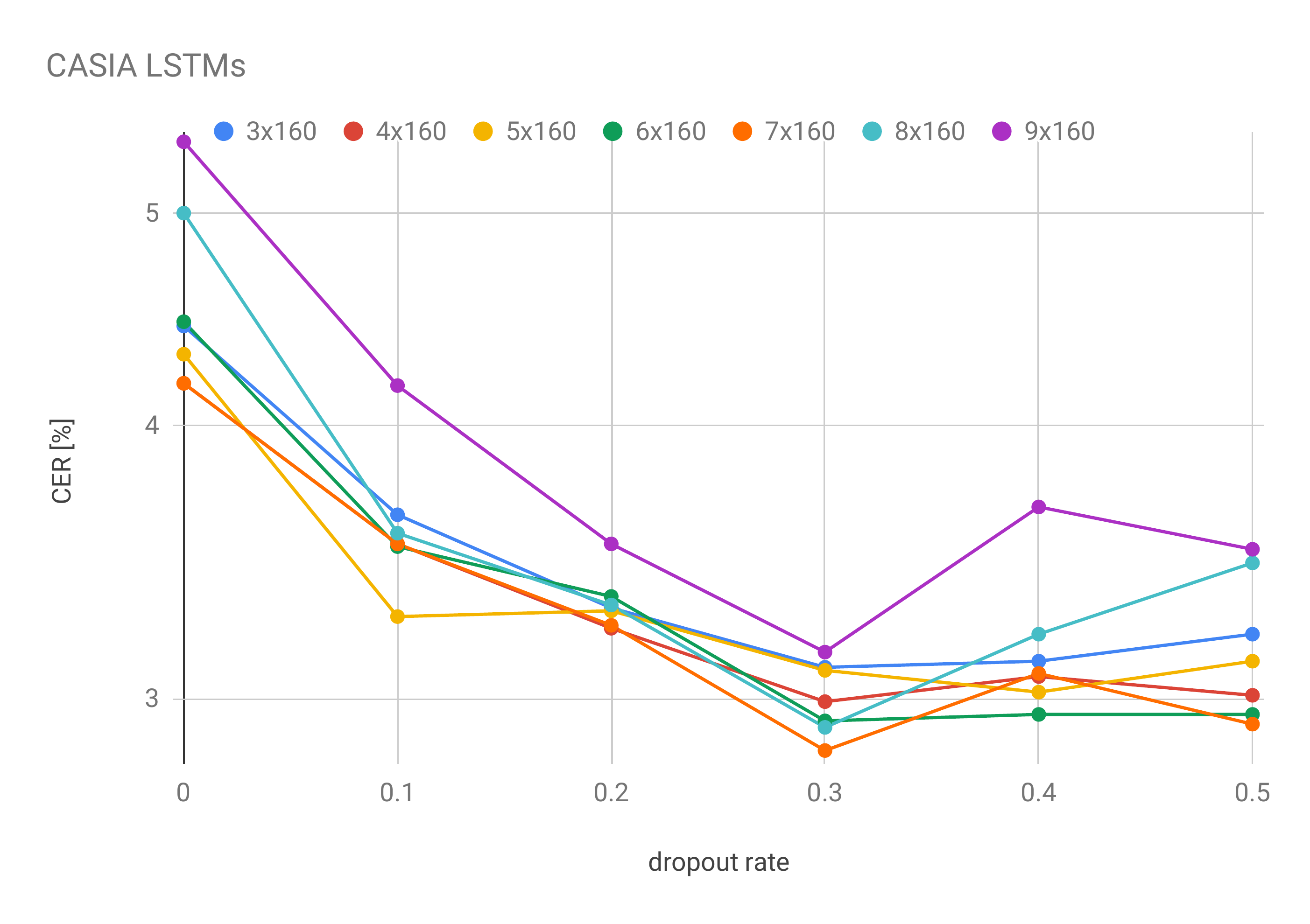}
    \caption{Character Error Rate (CER) over dropout rate as a function of model depth for the \iamondb and CASIA datasets using IndyLSTMs and LSTMs with a fixed layer width of 160 nodes. For both datasets and models, the optimal dropout rate is only weakly dependent on model depth.}
    \label{fig:dropout_depth}
\end{figure*}

Given the large number of experiments run over a wide range of models and dropout rates, we provide some additional data and analysis on the optimal dropout rate.
Figures~\ref{fig:dropout_width} and \ref{fig:dropout_depth} show the effect of the dropout rate on CER for both the \iamondb and CASIA datasets using both IndyLSTMs and LSTMs, as a function of model width and model depth, respectively.

Figure~\ref{fig:dropout_width} shows the CER as a function of the dropout rate for both IndyLSTMs and LSTMs trained on both the \iamondb and CASIA datasets, for models with a fixed depth of six layers and varying layer widths.
In all four setups, the optimal dropout rate increases with the layer width, e.g.\ for IndyLSTMs trained on the CASIA data, the optimal dropout rate is zero for the 6$\times$32 models, 0.1 for the 6$\times$64 models, and increases up to 0.4 with increasing layer width.
The range in which the optimal dropout rate varies is larger for the IndyLSTM models than for the LSTM models.

Analogously, Figure~\ref{fig:dropout_depth} shows the same results, yet varying the model depth using a fixed width of 160 nodes in each layer.
Although there is some indication that deeper models require higher dropout rates, the effect is restricted to 3--5 layers, and less obvious than those varying the width.

Finally, Figure~\ref{fig:latin_prod}, which varies both model width and depth, does not show a large variation in the optimal dropout rate, which varies between $0.2$ and $0.3$.

The results are not sufficiently consistent as to provide concrete guidance on how to select the optimal dropout rate, but the overall impact of the dropout rates do make clear that this is a hyperparameter that deserves some consideration and experimentation.

\end{document}